\documentclass[letterpaper, 10 pt, conference]{ieeeconf}  

\IEEEoverridecommandlockouts                              
\usepackage{graphicx}
\usepackage{subfigure}
\usepackage{lipsum}
\usepackage{cite}

\usepackage{fancyhdr}
\usepackage{color}
\usepackage{cancel,soul}
\usepackage{booktabs}
\usepackage{blindtext}

\usepackage{multirow}
\usepackage{gensymb}
\usepackage{makecell}
\usepackage{stix,bbding,pifont,fontawesome}
\usepackage{utfsym}
\usepackage{amsmath}
\usepackage{hyperref}
\graphicspath{
    {./fig/}
}
    
\title{\Large \bf OpenPneu: Compact Platform for Pneumatic Actuation with Multi-Channels} 

\author{Yingjun Tian, Renbo Su, Xilong Wang, Nur Banu Altin, Guoxin Fang, and~Charlie C. L. Wang$^{\dagger}$
\thanks{All authors are with Department of Mechanical, Aerospace, and Civil Engineering, The University of Manchester, United Kingdom}
\thanks{$^{\dagger}$Corresponding Author:~{\tt\footnotesize changling.wang@manchester.ac.uk}}
}

\usepackage{color}

\begin{document}

\maketitle

\begin{abstract}
This paper presents a compact system, OpenPneu, to support the pneumatic actuation for multi-chambers on soft robots. Micro-pumps are employed in the system to generate airflow and therefore no extra input as compressed air is needed. Our system conducts modular design to provide good scalability, which has been demonstrated on a prototype with ten air channels. Each air channel of OpenPneu is equipped with both the inflation and the deflation functions to provide a full range pressure supply from positive to negative with a maximal flow rate at 1.7 L/min. High precision closed-loop control of pressures has been built into our system to achieve stable and efficient dynamic performance in actuation. An open-source control interface and API in Python are provided. We also demonstrate the functionality of OpenPneu on three soft robotic systems with up to 10 chambers. 
\end{abstract}

\section{Introduction}\label{sec:Intro}
Soft robotics is an emerging topic in the robotics area due to its flexibility and adaptability to the complex 
environment\cite{softRobotDaniel}. As a feature of soft robot systems
, external power sources such as cable force, liquid, pressurized air
, and electronic voltage are generally applied to provide actuation~\cite{el2020soft}. This feature enables soft robots to achieve a high power-to-weight ratio~\cite{marchese2015recipe}. 
Throughout different types of actuation, pneumatic is environment-friendly and easy to use,
which is now widely applied in soft robotic research~\cite{joshi2021pneumatic, su2022pneumatic}. The safety and efficiency of pneumatic-driven soft robots have been proved in applications such as
fruit harvesting~\cite{navas2021soft} and minimally invasive surgeries~\cite{runciman2019soft}.

\begin{figure}[t]
\centering
\vspace{-5pt}
\includegraphics[width=\linewidth]{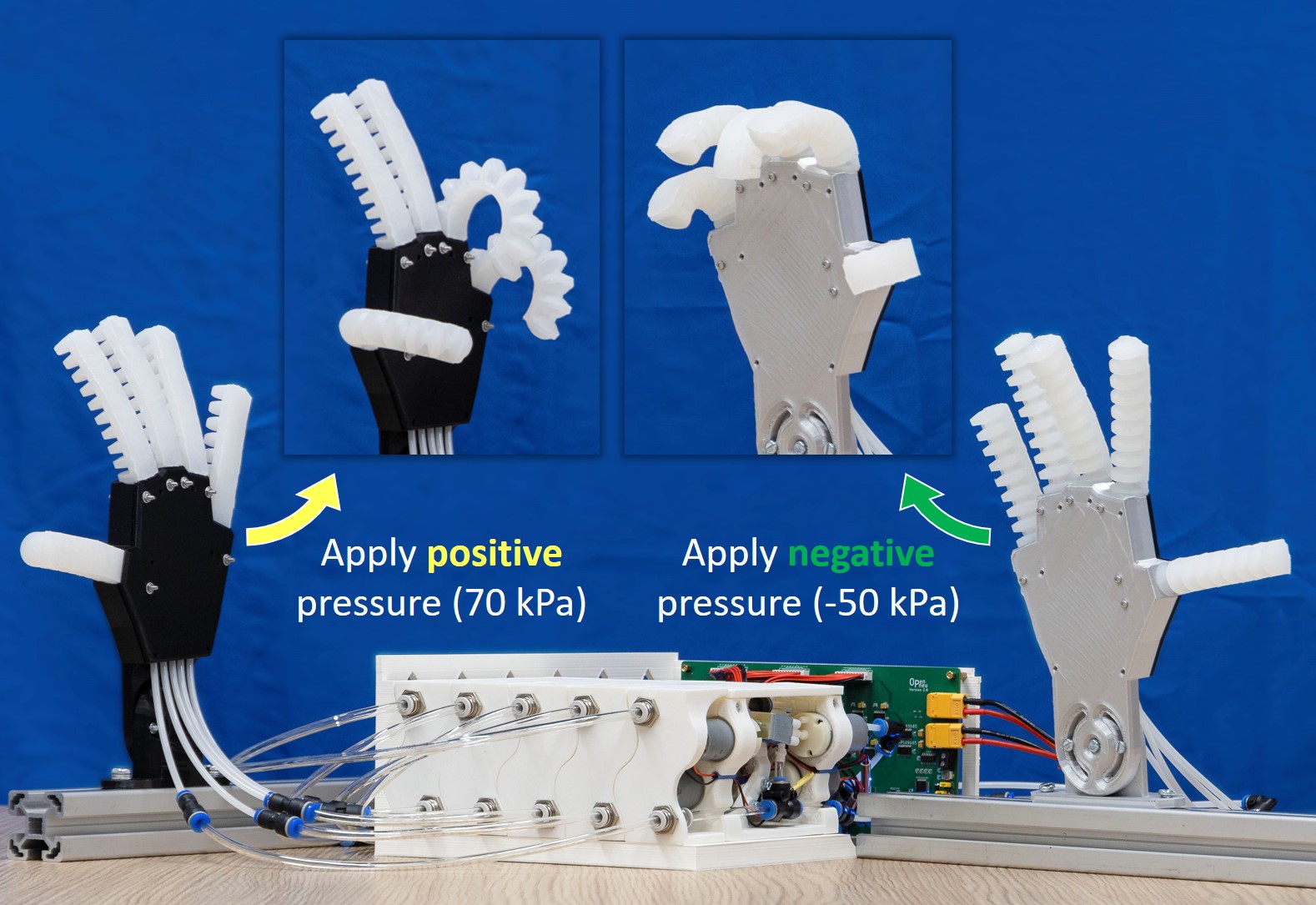}
\caption{
We present OpenPneu, an open-source platform with excellent mobility and scalability 
that can 
provide pressures for multi-channels at the same time. Both positive and negative pressures are realized, and the average response time is around one second when actuating two soft hands with ten fingers.
}\label{fig:teaser}
\vspace{-8pt}
\end{figure}

\begin{table*}[t] \scriptsize
\centering
\caption{Comparison of Different Pneumatic Actuation Systems}\label{tab:DifferentActSys}
\begin{tabular}{cc|c|c|c|c|c|c|c|c|c}         
\hline \hline                       
\specialrule{0em}{2pt}{1pt}         
\multicolumn{2}{c|}{\multirow{2}{*}{\textit{}~}}  & \multicolumn{9}{c}{Pneumatic Actuation Systems} \\ 
\specialrule{0em}{1pt}{2pt}         
\cline{3-11}                        
\specialrule{0em}{2pt}{1pt}         

\multicolumn{2}{c|}{\multirow{2}{*}{\textit{}~}}  
& \multicolumn{1}{c|}{Volume-based system} & \multicolumn{4}{c|}{Airflow-based system with compressed air}  & \multicolumn{4}{c}{Airflow-based system with micro-pump} 

\\ 

\specialrule{0em}{1pt}{2pt}         
\cline{3-11}  \specialrule{0em}{1pt}{2pt}
\multicolumn{2}{c|}{Property} & 
\multicolumn{1}{c|}{}&
\multicolumn{1}{c|}{Pneuduino}  & 

\multicolumn{1}{c|}{PneumaticBox} &
\multicolumn{1}{c|}{Zhou \textit{et al.}}&
\multicolumn{1}{c|}{Festo}&
\multicolumn{1}{c|}{SoRoToolkit}& %
\multicolumn{1}{c|}{Zhang \textit{et al.}}&
\multicolumn{1}{c|}{FlowIO}&
\multicolumn{1}{c}{OpenPneu}

\\ 
\specialrule{0em}{1pt}{2pt}
\hline \specialrule{0em}{1pt}{2pt}
\multicolumn{2}{c|}{Source}  & ~\cite{RusIJRR,TolleySyringe,SoRoLearning_Syringe,mannequin_Syringe} & ~\cite{pneuduino} & \cite{TuBerlinSystem}  & ~\cite{BCL13} & \cite{festoMotionTerminal} & \cite{softroboticstoolkit}& \cite{HongenPneu}& \cite{FlowIO}& Ours\\
\multicolumn{2}{c|}{Airway numbers$^\dagger$} & up-to 16~\cite{RusIJRR}  & 1 & 8  & 13 & 10 & 1 & 6 & 1 & 10\\
\multicolumn{2}{c|}{High Pressure Output$^\star$} & \ding{55} & \ding{51} & \ding{51} & \ding{51} & \ding{51} &  \ding{55} & \ding{55} &  \ding{55}&  \ding{55}\\

\multicolumn{2}{c|}{Support Vacuum} & \ding{51} & \ding{55}  & \ding{55} & \ding{55} & \ding{51} & \ding{55} & \ding{51} & \ding{51}& \ding{51}\\
\multicolumn{2}{c|}{Free of External$^\ddagger$} & \ding{51}    & \ding{55}  & \ding{55}  & \ding{55} & \ding{55} & \ding{51} & \ding{51}& \ding{51}& \ding{51}\\ 
\multicolumn{2}{c|}{Compactness}&  \ding{55} & \ding{51} & \ding{51}     & \ding{55} & \ding{51}& \ding{51}&  \ding{55}& \ding{51}& \ding{51} \\
\multicolumn{2}{c|}{Open Source}& \ding{55}  & \ding{51} & \ding{51} & \ding{55}
& \ding{55} & \ding{51} & \ding{55} & \ding{51} & \ding{51}\\ 
\multicolumn{2}{c|}{Rough Cost$^\sharp$ (\$)}&  $\geq$1000 & N/A & 2020 & 830
& 6600
& 510
& 780
& 2650 &540\\
\specialrule{0em}{0pt}{2pt} \hline\hline
\end{tabular}
\begin{flushleft}
$^\dagger$Independent number of air channels that can be handled simultaneously by the actuation system.
$^\ddagger$Indicates whether the actuation system needs the external pressure source. \\
$^\star$Defined by pressure higher than 2bar. \quad \ \ $^\sharp$The cost used to build ten independent air channels.
\end{flushleft}\label{tab:Comparison}
\vspace{-10pt}
\end{table*}

While novel designs and control strategies for pneumatic-driven soft robots have been sufficiently studied~\cite{softRobotDaniel, mosadegh2014pneumatic,fang2020kinematics}, their performance also relies on the development of an efficient
actuation platform that can effectively drive chambers to the required pressures~\cite{joshi2021pneumatic}. As an example, to drive two soft robotic hands 
to the gestures of the rock-paper-scissors game as illustrated in Fig.~\ref{fig:teaser}, the actuation system needs to rapidly and simultaneously change the pressures in 
ten air chambers. On the other hand, we see a trend of soft robot designs that require both positive and negative pressures to achieve better functionalities (e.g., to obtain larger workspace~\cite{TolleySyringe} or variable stiffness~\cite{xu2020soft}). To support the development of novel pneumatic-driven soft robots, 
an actuation system should have the following properties:
\begin{enumerate}
    \item A modular design to realize good scalability for soft robots with multiple chambers.
    \item Ability to provide positive/negative pressures as output with stable and efficient dynamic performance.
    \item Ease to be reproduced with a user-friendly programming interface to support soft robotic applications.
\end{enumerate}
Simultaneously achieving all these properties on the same actuation system has not been found in prior research.
Pneuduino~\cite{pneuduino} and PneumaticBox~\cite{TuBerlinSystem} are two integrated solutions with high-pressure compressed airflow as the input. Soft Robotics Toolkit~\cite{softroboticstoolkit}, is another well-known low-cost solution built with off-the-shelf components, however, have limited scalability for multi-channel pressure supply. FlowIO~\cite{FlowIO} developed by Ali Shtarbanov has good portability, however only supports rapid pressure change in one channel at one time. 
A review of existing systems with a comprehensive comparison of their properties is presented in Sec.~\ref{subSec:relatedWork} and Table~\ref{tab:DifferentActSys}.
Nevertheless, fewer studies have been reported focused on developing general and open-sourced pneumatic actuation systems. We see researchers in this field generally build their own actuation devices from the scratch 
with a time-consuming building process (e.g., \cite{TolleySyringe,SoRoLearning_Syringe,mannequin_Syringe}). 

In this paper, we present a self-contained pneumatic actuation platform named OpenPneu that can provide all the aforementioned properties
to drive soft robots. OpenPneu is with modular design to support multi-channel output driven by micro-pumps, where each channel has the ability to provide both positive and negative pressures 
ranging between $\left[-50, 80\right]$ kPa. 
Specially designed mechanical structure and customized PCBs are built to allow good scalability of the system. We verify the effectiveness of OpenPneu on various soft robots fabricated by
silicon rubbers~\cite{sun2013characterization, mosadegh2014pneumatic}, with a presented example prototype that supports up to ten chambers being actuated simultaneously. The static and dynamic performance of the system are tested, and the ability to provide pressure compensation under external disturbance is demonstrated. Important data including the mechanical and circuit design, the firmware program, 
the instruction set and the coding interface are made publicly accessible at 
\url{https://openpneu.github.io/OpenPneu}.

\begin{figure}[t]
\centering
\vspace{-0pt}
\includegraphics[width=1.0\linewidth]{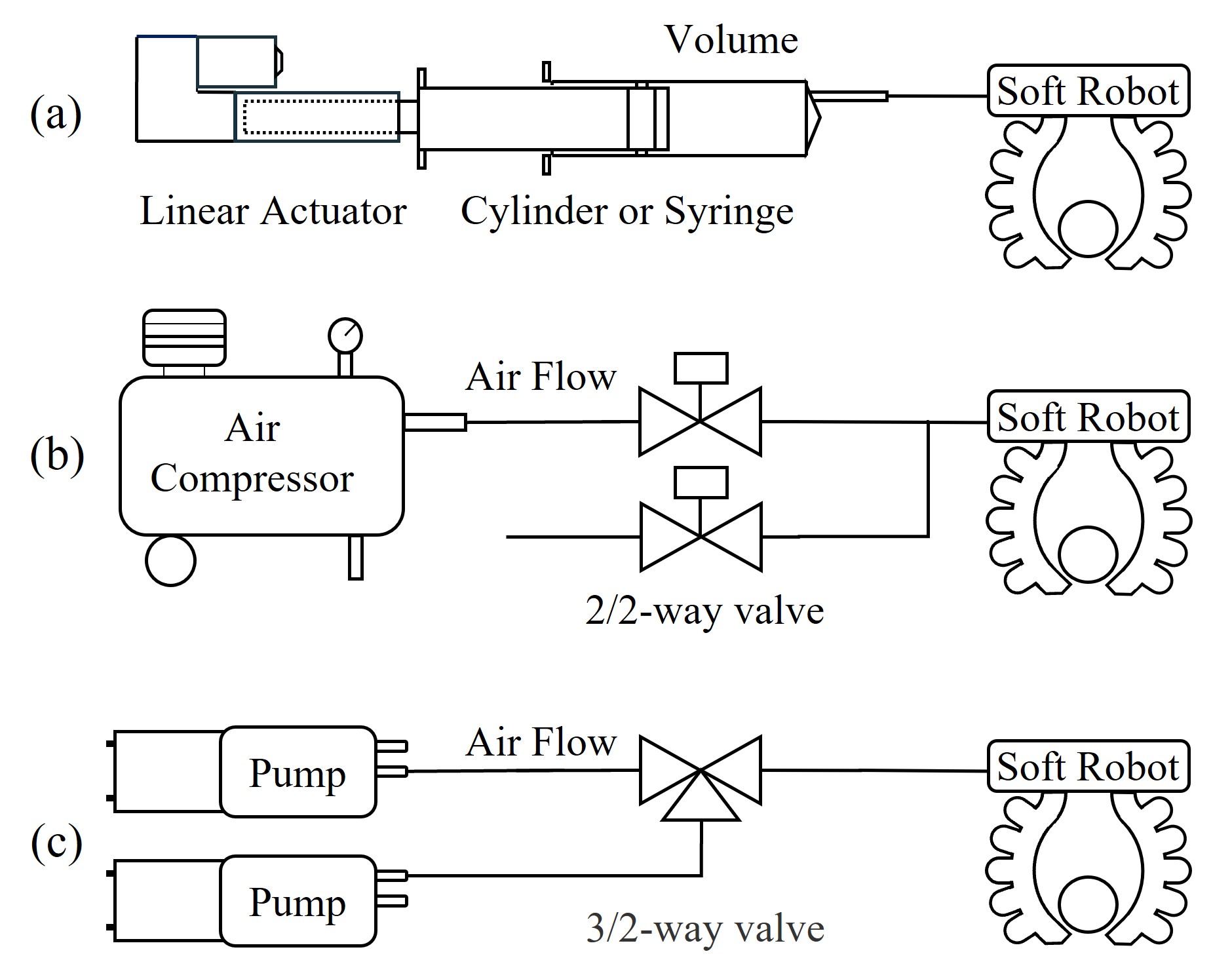}\\
\caption{
Working principles of different pneumatic systems developed for soft robotic applications: (a) Volume-based actuation system, (b) System based on an external supply of compressed air,
and (c) Our self-contained system with micro-pumps to generate airflow with controlled pressure. 
}\label{fig:sysDiff}
\end{figure} 

\section{Related Work}
\label{subSec:relatedWork}


Existing pneumatic actuation solutions can be categorized as volume-based and airflow-based systems. An illustration of the working principles is shown in Fig.~\ref{fig:sysDiff}.

\subsection{Volume-Based System}
\label{subSec:volumeBased}
The volume-based systems are widely used in soft robotics research~\cite{TolleySyringe,SoRoLearning_Syringe,mannequin_Syringe} since they are easy to control and able to provide continuous pressure changes. As shown in Fig.~\ref{fig:sysDiff}(a), the pressure supplement is usually 
driven by the volume variation of cylinders or syringes and controlled by the motion of linear actuators. The volume-based 
actuation systems have good scalability in both mechanical and control design. Marchese \textit{et al.}~\cite{RusIJRR} have demonstrated
their ability to effectively drive a soft manipulator with 16 chambers by volume-based actuation. However, those systems have limited mobility
since they are constructed with heavy linear actuators and large syringes with nearly the same volume as the maximally inflated chamber.
Meanwhile, as a closed system without compensated airflow, it can lose designed functionalities after driving the system with large cycles~\cite{mannequin_Syringe}. This is the case when chambers have air-leakage issues, which has less influence on our airflow-based system (a comparison is shown in Sec.~\ref{subsec:mannequin}).

\subsection{Airflow-Based System with Compressed Air}
When the soft robot systems require working pressures higher than 2 bar,
airflow-based systems with an external supply of compressed air 
are generally used. These systems usually contain a
proportioning valve~\cite{festoMotionTerminal} (e.g., proportional-pressure regulators from FESTO or SMC) or a high-frequency electromagnetic valve with a personalized controller~\cite{BCL13,pneuduino,TuBerlinSystem} to realize fast and efficient pressure control. 
However, with the nature of using compressed air as input, these systems are hard to provide negative pressure and have poor mobility.

\subsection{Airflow-Based System with Micro-pump}

As a self-contained system, the solution that uses micro-pumps to provide airflow 
has the ability to generate both positive and negative pressures as output. 
Based on this design, 
researchers build cost-efficient systems with off-the-shelf components~\cite{softroboticstoolkit, HongenPneu, programmableair}. An open-source system named FlowIO with a highly integrated design is presented in~\cite{FlowIO}, which is now being used in many soft robotic applications. 
However, it can only sequentially adjust the pressure change on various air channels, which limits its usage in parallel soft robot systems that require several chambers to be actuated together~\cite{TuBerlinSystem}. Instead, the OpenPneu prototype demonstrated in this work shows its ability to provide up to ten pressure outputs simultaneously with a fast response performance. 

\section{Hardware}\label{secHardware} 

\begin{figure}[t]
\centering
\includegraphics[width=1\linewidth]{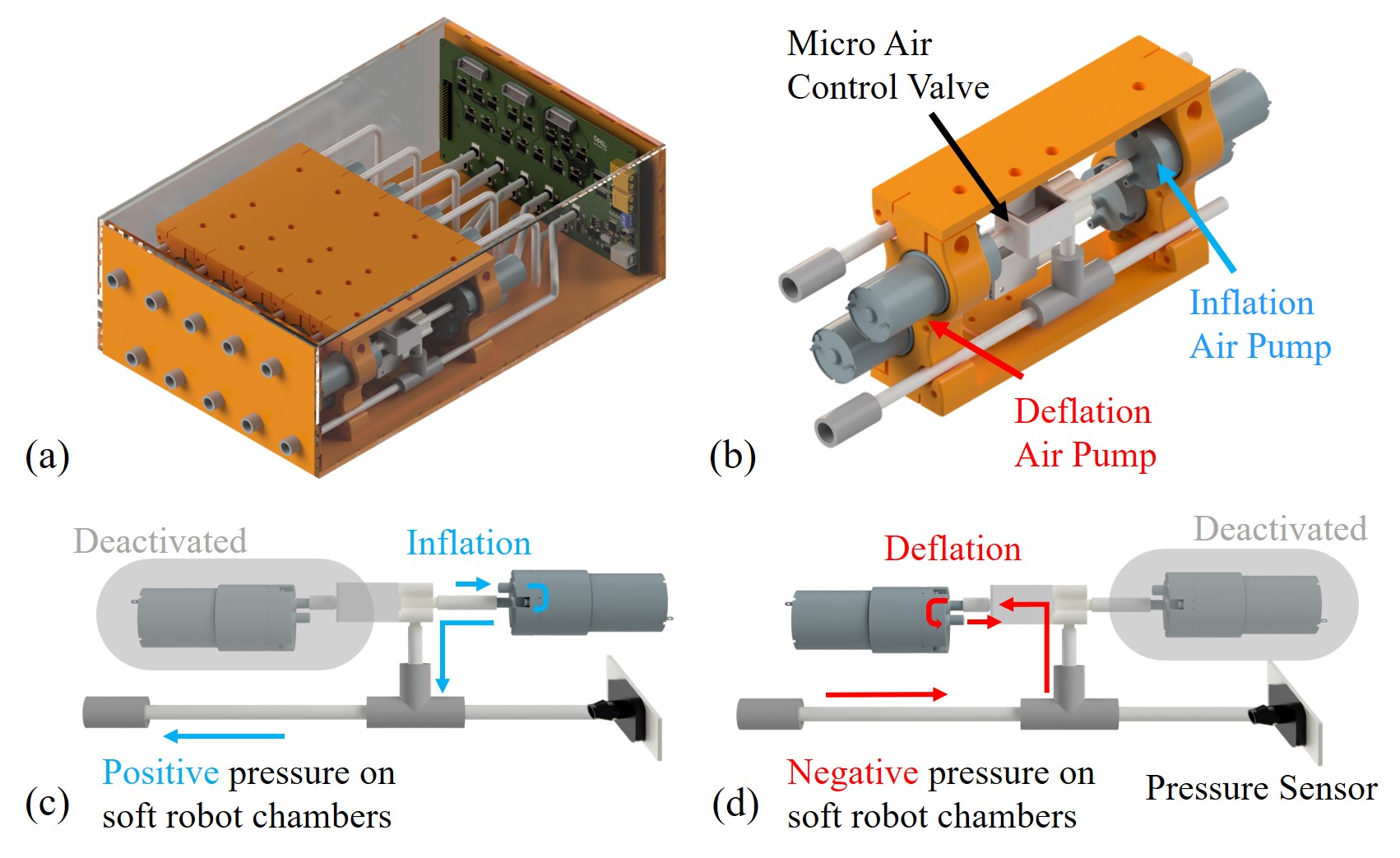}\\
\caption{(a) Hardware of our OpenPneu platform. (b) A single module contains two output channels driven by four air pumps and two
air valves. (c, d) Illustration of pressure regularization and the working principles of inflation and deflation respectively.
}\label{fig:systemIntro}
\end{figure}

We first present the hardware of OpenPneu which contains a set of compact modules for pressure regularization and a highly integrated circuit board. 
The assembly drawing of our hardware can be found in Fig.~\ref{fig:systemIntro}(a).

\subsection{Mechanical Design}
To achieve good scalability and compactness in the mechanical part, we designed each module with a pair of two airflow channels. Each channel is equipped with two micro-pumps and one valve. The working principles are illustrated in Fig.~\ref{fig:systemIntro}(c)-(d), where one micro-pump provides inflation and the other one works for deflation to generate positive/negative pressures, respectively.
The valve is controlled to switch between working 
functions and only allows one side of the airflow to go into the chambers of soft robots. Pressure sensors are connected on the other side to monitor 
pressure levels in air channels. The mechanical fixture of each module is designed with an 'S' shape on both sides, which provides the self-interlocking between modules and allow an easy assembly/disassembly process to support good scalability.

\subsection{Circuit Design}

\begin{table}[t] \scriptsize
\centering
\caption{Bill of Materials$^\dagger$
}
\label{tab:BoM}

\begin{tabular}{c|c|c|c}         
\hline \hline                       
\specialrule{0em}{2pt}{1pt}         
\multicolumn{1}{c|}{Component} & \multicolumn{1}{c|}{Cost per item} & \multicolumn{1}{c|}{Quantity} & \multicolumn{1}{c}{Total cost} \\
\specialrule{0em}{1pt}{2pt}         
\cline{1-4}                        

\specialrule{0em}{2pt}{1pt}         
\multicolumn{1}{c|}{Microcontroller} & \multicolumn{1}{c|}{\$3.4} & \multicolumn{1}{c|}{1}
& \multicolumn{1}{c}{\$3.4} \\
\specialrule{0em}{1pt}{2pt}    
\cline{1-4}                        

\specialrule{0em}{2pt}{1pt}         
 \multicolumn{1}{c|}{I/O Expansion Chip} & \multicolumn{1}{c|}{\$1.3} & \multicolumn{1}{c|}{1} & \multicolumn{1}{c}{\$1.3}\\
\specialrule{0em}{1pt}{2pt}   
\cline{1-4}                        

\specialrule{0em}{2pt}{1pt}         
\multicolumn{1}{c|}{Positive Pressure Sensor(0, 15 psi)} & \multicolumn{1}{c|}{\$20.7} & \multicolumn{1}{c|}{5} & \multicolumn{1}{c}{\$103.5}\\
\specialrule{0em}{1pt}{2pt}   
\cline{1-4}                        

\specialrule{0em}{2pt}{1pt}         
\multicolumn{1}{c|}{Hybrid Pressure Sensor($\pm$15 psi)} & \multicolumn{1}{c|}{\$19.6} & \multicolumn{1}{c|}{5} & \multicolumn{1}{c}{\$98.0}\\
\specialrule{0em}{1pt}{2pt}   
\cline{1-4}                        

\specialrule{0em}{2pt}{1pt}         
 \multicolumn{1}{c|}{PWM Generator} & \multicolumn{1}{c|}{\$2.7} & \multicolumn{1}{c|}{2} & \multicolumn{1}{c}{\$5.4}\\
\specialrule{0em}{1pt}{2pt}   
\cline{1-4}                        

\specialrule{0em}{2pt}{1pt}         
 \multicolumn{1}{c|}{Pump \& Valve Driver} & \multicolumn{1}{c|}{\$4.0} & \multicolumn{1}{c|}{15} & \multicolumn{1}{c}{\$60.0}\\
\specialrule{0em}{1pt}{2pt}   
\cline{1-4}                        


\specialrule{0em}{2pt}{1pt}         
\multicolumn{1}{c|}{Micro Air Pump} & \multicolumn{1}{c|}{\$6.0} & \multicolumn{1}{c|}{20} & \multicolumn{1}{c}{\$120.0}\\
\specialrule{0em}{1pt}{2pt}    
\cline{1-4}                        

\specialrule{0em}{2pt}{1pt}         
 \multicolumn{1}{c|}{Micro Air Valve} & \multicolumn{1}{c|}{\$3.0} & \multicolumn{1}{c|}{10} & \multicolumn{1}{c}{\$30.0}\\
\specialrule{0em}{1pt}{2pt}               
\cline{1-4}                        

\specialrule{0em}{2pt}{1pt}         
 \multicolumn{1}{c|}{Printed Circuit Board} & \multicolumn{1}{c|}{\$63.7} & \multicolumn{1}{c|}{1} & \multicolumn{1}{c}{\$63.7}\\
\specialrule{0em}{1pt}{2pt}               
\cline{1-4}                        

\specialrule{0em}{2pt}{1pt}         
 \multicolumn{1}{c|}{3D Printed Components} & \multicolumn{1}{c|}{ } & \multicolumn{1}{c|}{6} & \multicolumn{1}{c}{\$60.0}\\
\specialrule{0em}{1pt}{2pt}               
\cline{1-4}                        

\specialrule{0em}{2pt}{1pt}         
\multicolumn{1}{c|}{Total} &\multicolumn{2}{c}{}
&\multicolumn{1}{c}{\$545.3}\\

\hline\hline

\end{tabular}
\begin{flushleft}
$^\dagger$Detailed information for electronic components and drawings of 3D printed models can be found on the page provided in Sec.~\ref{sec:Intro}.
\end{flushleft}

\vspace{-5pt}
\end{table}

\begin{figure*}[t]
\centering
\includegraphics[width=1\linewidth]{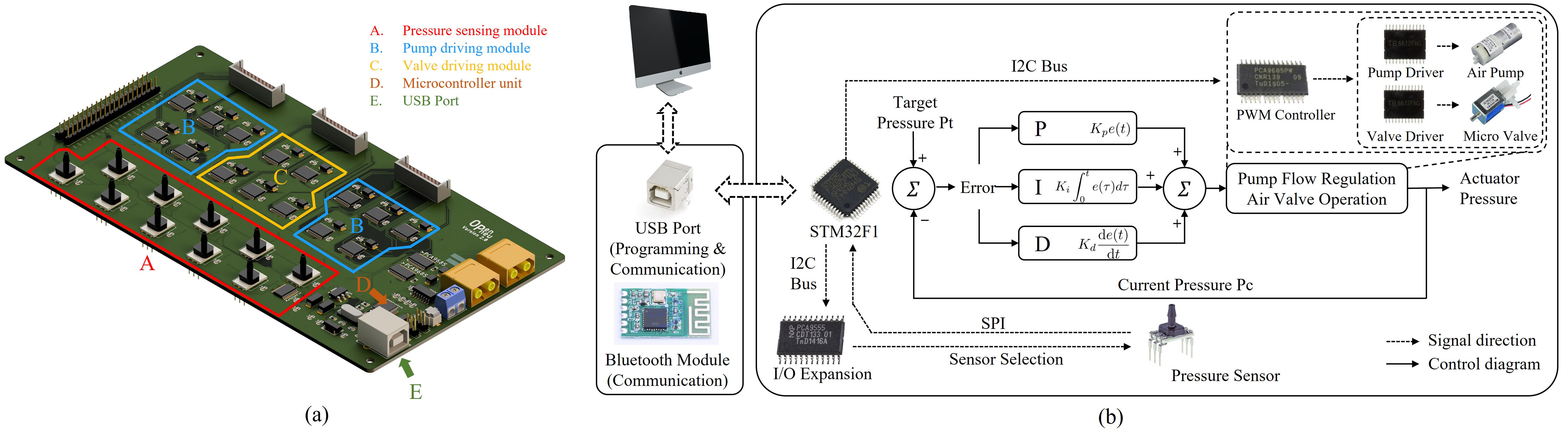}\\
\vspace{-5pt}
\caption{(a) PCB design for OpenPneu with integrated sensing and driving modules. (b) Illustration of firmware architecture: The main loop of MCU is designed for communication while the PID-based pressure controller is operated in 
the timer interrupt function.
}\label{fig:firmware_PID}
\end{figure*} 

Since the OpenPneu is designed to support pressure supply for multiple channels, 
the circuit board should be able to provide multiple signals to control valves and pumps under high power load especially when all the channels are being operated. 
As there is no existing commercial \textit{Printed Circuit Board} (PCB) solution available for integrated pressure sensing and regulation for multiple air channels, a new PCB has been personalized to meet all requirements.

The PCB of OpenPneu is illustrated in Fig.~\ref{fig:firmware_PID}(a) which contains various modules. STM32F103 is selected as the microcontroller (MCU) for operating all input commands. Information on the main components used to build the system can be found in Table. \ref{tab:BoM}. One of the main modules is for sensing the pressure, which contains a set of pressure sensors connected with an I/O expansion chip acting as a pressure selector 
to efficiently provide information to the MCU. Note that the end-user is able to change
the type of pressure sensors (e.g., pressure range and precision) based on different applications. For the demonstration purpose, we select five high-precision positive sensors and another five sensors that have the ability to detect both positive and negative pressures. The other two main modules are used to drive the pumps and valves, 
where both are equipped with pulse-width modulation (PWM) drivers to generate adjustable independent 16 PWM signals at a resolution of 4096 bits. The main circuit of the PCB board is designed to withstand a maximal working current at 20A, which is sufficient enough for an extension of 12 modules (i.e., 24 air channels). 


\vspace{5px}

Our system can achieve smooth control of pressure output with the range of $\left[-50, 80\right]$ kPa, which fundamentally resolves the air-leaking problem in the soft robot system (see the experiment shown in Sec.~\ref{subsec:mannequin}). As the modular design has been considered for both the mechanical and the electrical aspects
of the system, users are able to customize the number of air channels
and the range of pressure output to meet the needs of particular applications. The total cost of the prototype we demonstrated in this work with 10 channels is less than $\$$550 (see the summary in Table~\ref{tab:BoM}), where
60\% of the total cost from the pressure sensors and the micro air pumps are used to realize high-precision pressure control. 
This cost is less than most of the existing solutions based on the comparison presented in Table~\ref{tab:Comparison}.

\section{Software}\label{secSoftware}
The control architecture of OpenPneu is presented in this section, which contains low-level firmware and high-level \textit{Application Programming Interface} (API) functions.
These together support an effective driver and communication with the circuit board to control pumps and valves. A Python-based coding interface are created to guarantee a full access to the system functionalities, which support the development of soft robotic applications.


\begin{figure}[t]
\centering
\includegraphics[width=1.0\linewidth]{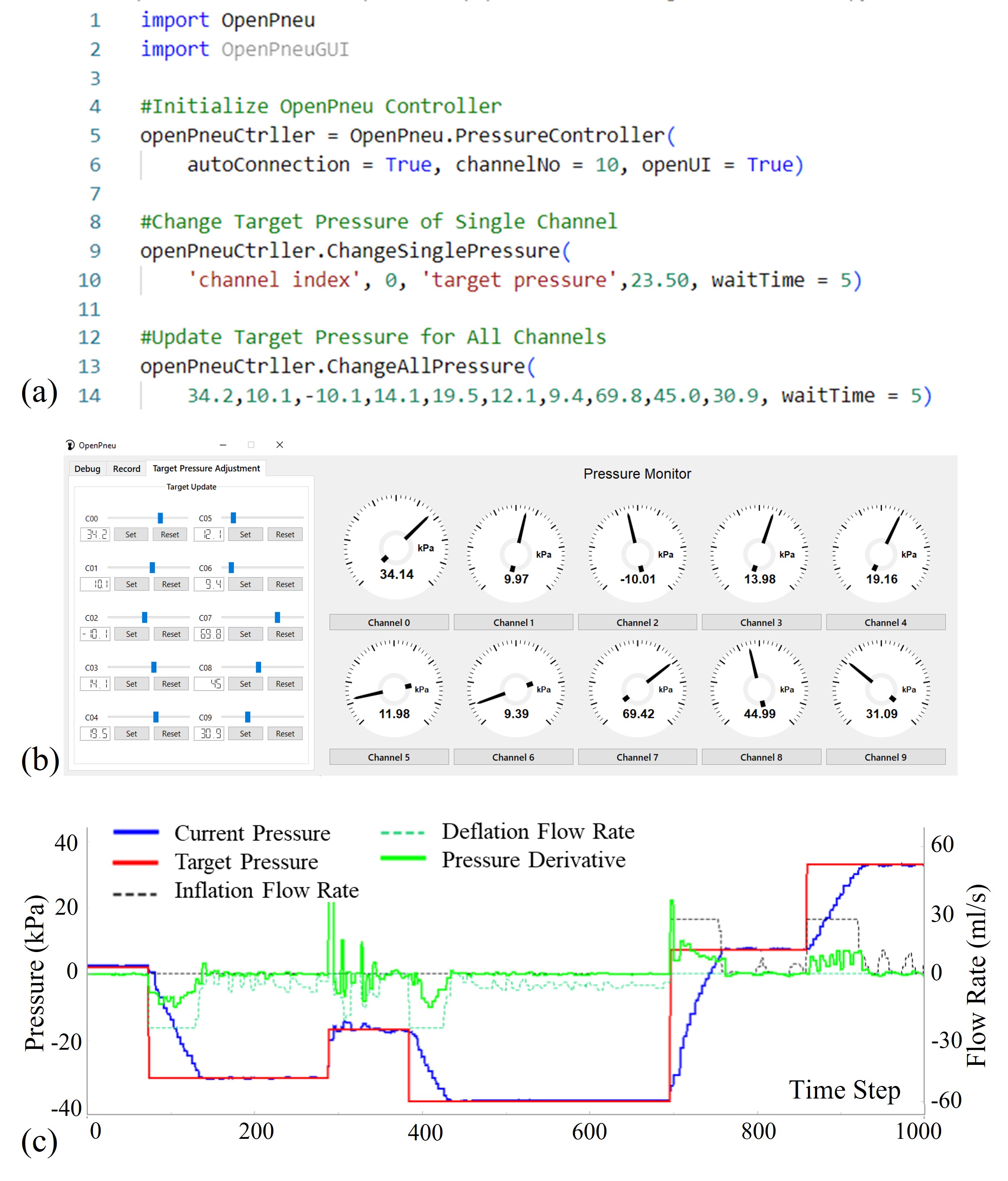}\\
\vspace{-5pt}
\caption{(a) Example Python code to control the pressure output of OpenPneu system. (b) An interactive user interface is provided to direct adjust the pressure on each channel. (c) The pressure and flow rate can be visualized in real-time.
}\label{fig:GUI}
\vspace{-10pt}
\end{figure} 

\subsection{Firmware and Low-level Pressure Control}
The firmware of OpenPneu is illustrated in Fig.~\ref{fig:firmware_PID}(b) and developed within Arduino IDE based on STM32duino\cite{STM32duino}. Pressure control for each channel 
is achieved by using the \textit{proportional–integral–derivative} (PID) controller with sensor signal as feedback. The target pressure is taken as input and commands are sent to MCU through the USB port or Bluetooth module via serial communication. MCU computes the pressure difference to the target and sends signals to speed controllers which generate PWM waves to guide pump and valve drivers with desired voltages. In our implementation, the PID controller is run in timer interrupt function on MCU rather than in a computer program to achieve fast and stable pressure regularization. The frequency of MCU is at $50$Hz.

\subsection{Coding Interface}
In our system, the instruction set is built as cross-platform 
commands to control the OpenPneu system through serial communication. To provide a simple and user-friendly coding interface for researchers, both the \textit{application programming interface} (API) and the \textit{graphical user interface} (GUI) are provided. Example Python code of using API to individually control a single channel or all channels together is demonstrated in Fig.~\ref{fig:GUI}(a). The GUI shown in Fig.~\ref{fig:GUI}(b) is based on PyQt5, which allows users to interactively adjust target pressure in each channel. As shown in Fig.~\ref{fig:GUI}(c), the current pressure and airflow rate could be visualized in the selected channels. All functions supported by our coding interface can be found in the documentation on GitHub.

\begin{figure}[t]
\centering
\includegraphics[width=0.9\linewidth]{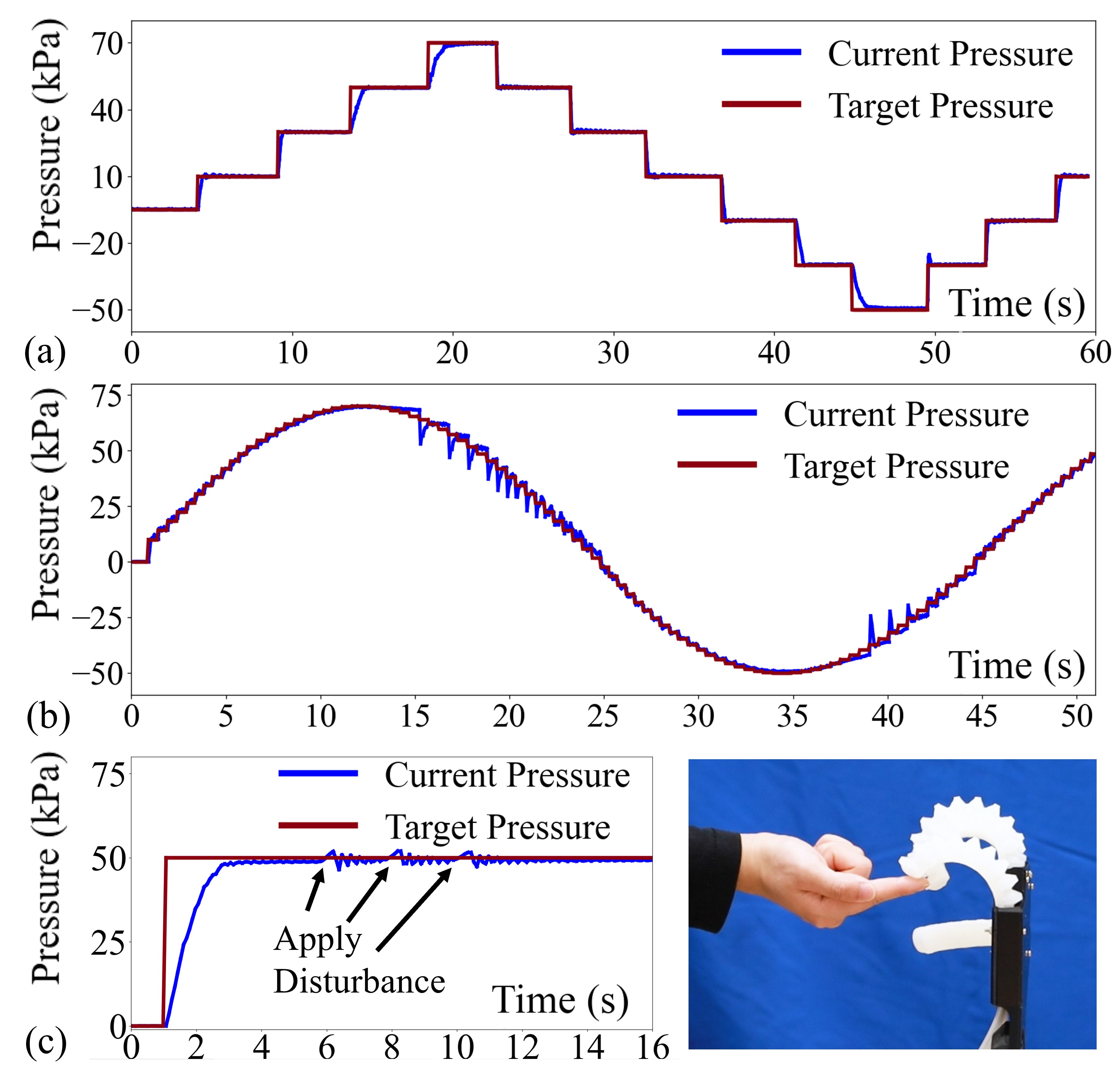}\\
\caption{The performance of the proposed OpenPneu system has been tested on a soft robotic finger with bi-directional deformations. (a) Step response for pressure control. (b) Response analysis for continuous target pressure variation. 
(c) Robustness to handle disturbance on the actuated chamber.
}\label{fig:closedloopPerfomance}
\end{figure}

\begin{figure*}[t]
\centering
\includegraphics[width=1.0\linewidth]{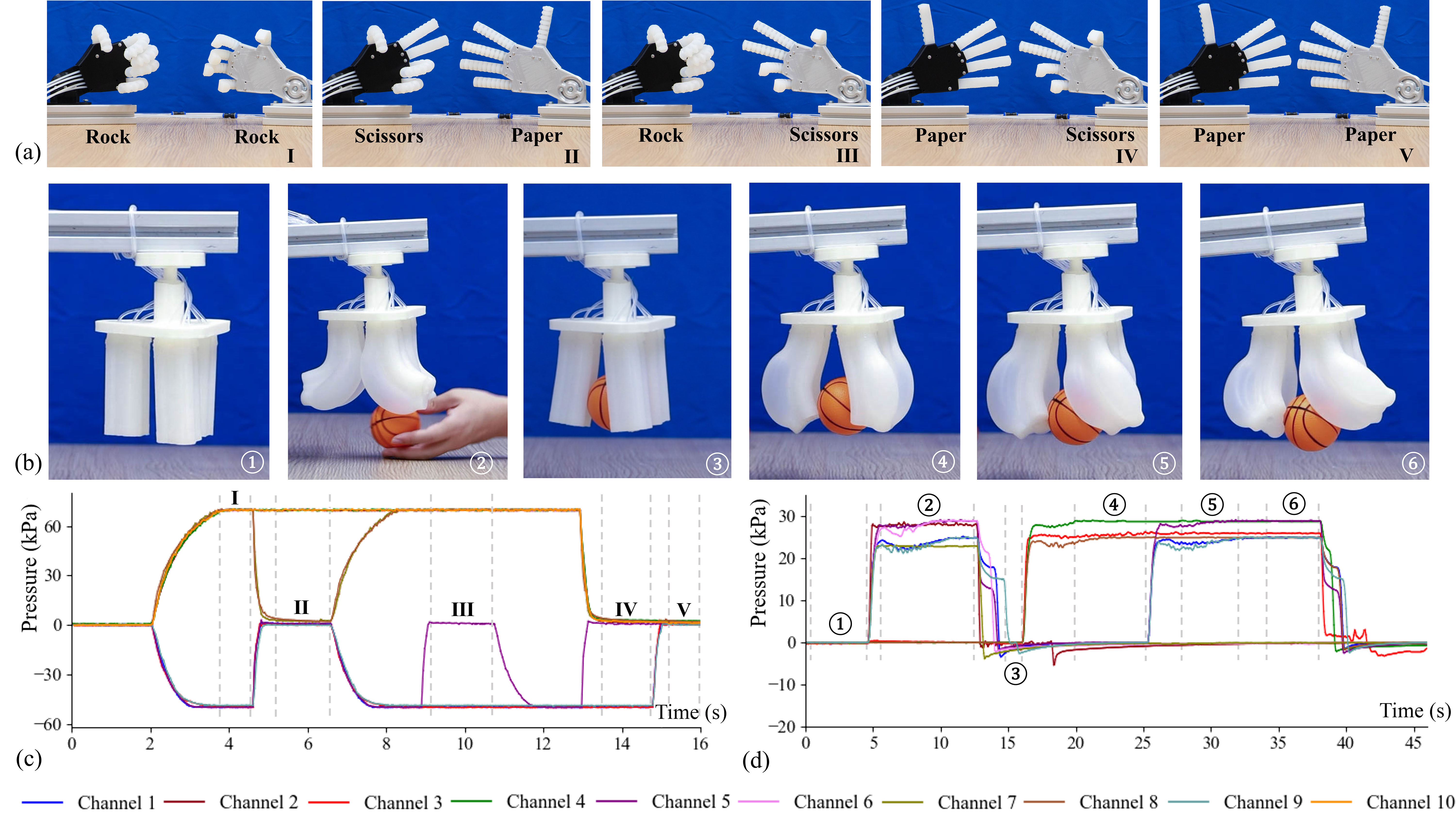}\\
\vspace{-5pt}
\caption{
Applying our OpenPneu system to drive soft robots with multiple chambers. (a) Two soft robotic hands are actuated to play the ``Rock-Paper-Scissors'' game. (b) A soft gripper with nine chambers is actuated to grasp and twist a small basketball. (c, d) Visualization of the pressure change in channels to demonstrate the dynamic performance of our system -- (left) soft hands with channels 1-10 and (right) the soft gripper with channels 1-9. 
}\label{fig:sequence}
\vspace{-15pt}
\end{figure*} 

\begin{figure}[t]
\centering
\includegraphics[width=1.0\linewidth]{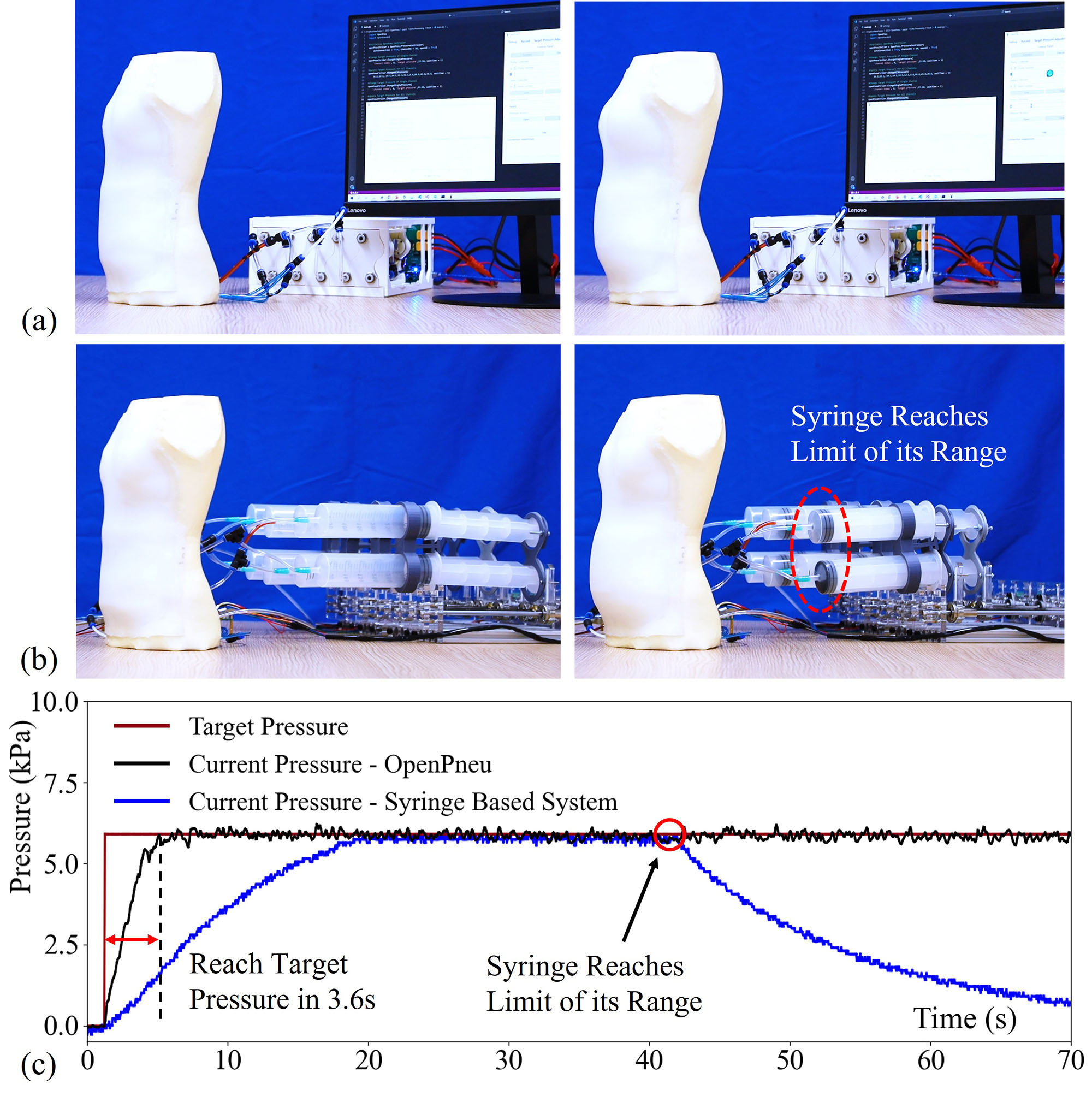}\\
\vspace{-10pt}
\caption{Behaviour comparison on actuating soft robotic mannequin with (a) our OpenPneu platform and (b) a volume-based actuation system~\cite{mannequin_Syringe}. (c) Visualization of the pressure change on a chamber with the air-leaking issue. The right figure shows the zoom-in on times sequence from 0-20 sec. 
}\label{fig:mannequin}
\vspace{-5pt}
\end{figure} 

\section{Case studies and Applications}\label{secResultApplication}

We have implemented the proposed 
hardware and software, and a prototype of OpenPneu is shown in Fig.~\ref{fig:teaser}. The dimensions of the system equipped with ten air channels are $360 \times 230 \times 120$ mm. All customized mechanical components are fabricated by 3D printing 
and PCBs are fabricated through JLC-PCB company with surface mount technology. The assembly time of the whole system is around 3 hours 
for all parts. To test the performance and verify the usage of OpenPneu in practical applications, we conduct three case studies with different soft robotic 
setups. We quantitatively tested both static and dynamic performance of the system when providing positive and negative pressures. 
The performance of the OpenPneu on soft robot applications can also be found in the supplemental video of this paper.

\subsection{Soft Robotic Hands}~\label{subsec:softhands}
We first demonstrate the performance of OpenPneu on two soft robot hands 
with ten fingers. Each finger is designed with teeth-like structures~\cite{mosadegh2014pneumatic} and can be actuated by both positive and negative pressures. The finger is fabricated by Smooth-On Dragon Skin 30 silicon rubber and can bend bi-directionally to 120 degrees.
The step response and continuous pressure tracking performance of our system are tested, with the result plotted in Fig.~\ref{fig:closedloopPerfomance}(a) and Fig.~\ref{fig:closedloopPerfomance}(b), respectively.
Our platform shows good stability for both tasks, with an average settling time of 0.9 second and a steady-state error of 0.07 kPa. 
This response time is fast enough to rapidly drive soft robotic system~\cite{mosadegh2014pneumatic}. When the target of the system is changing frequently, OpenPneu can also effectively control the airflow to reach the target pressure.
However, the system becomes less stable when the pressure continuously goes down (see the portion with oscillation on the curve in Fig.~\ref{fig:closedloopPerfomance}(b)). This is because the micro-valve can only switch between discrete states and show less ability to discharge the airflow.
We also test the case when the disturbance is applied to the soft finger (i.e., human interaction applied as shown in Fig.~\ref{fig:closedloopPerfomance}(c)). The system takes an average of 1.5 seconds to recover. Note that the silicon rubber used to fabricate soft fingers is extremely soft and with a low stiffness and damping coefficient.

We also test the performance difference between individual modules when driving ten soft fingers simultaneously with both positive and negative target pressure. As shown in Fig.~\ref{fig:sequence}(a),
soft hands are controlled to play the 'Rock Paper Scissors' game and are able to change the configuration around one second. 
Also, we could see the pressure curves in different chambers overlapped with each other in Fig.~\ref{fig:sequence}(c), and therefore shows nearly the same performance.

\subsection{Soft Robotic Gripper}
The application of using OpenPneu to drive a soft robotic gripper with multi-chambers
is presented in Fig.~\ref{fig:sequence}(b). The soft gripper consists of three manipulators fabricated by Smooth-On Ecoflex 0030 silicon rubber. Each manipulator has three parallel chambers and is able to deform in 3D space~\cite{marchese2016design}. The pressure changes in all nine channels are visualized in Fig.~\ref{fig:sequence}(d). It can be observed that different modules are well cooperated with each other to drive the soft gripper to grasp and rotate a small basketball with maximum pressure at 30 kPa. The average response time for all actions is 0.9 seconds. When the gripper contact with the object, the pressure values inside the chambers will change and our system can still successfully handle the interaction. 
The time used to stabilize the system is at average 1.6 seconds.
\subsection{Soft Robotic Mannequin} \label{subsec:mannequin}
In the third case study, OpenPneu is used to actuate a soft robotic mannequin~\cite{mannequin_Syringe} consisting of four chambers as illustrated in Fig.\ref{fig:mannequin}. Both the OpenPneu-based (Fig.~\ref{fig:mannequin}(a) and the syringe-based systems (Fig.~\ref{fig:mannequin}(b)) are tested to deform the mannequin into a target shape according to the given shape of a human body. It can be found that our system can only take $20\%$ of the time (3.6 sec. v.s. 18.0 sec.) as shown in Fig.~\ref{fig:mannequin}(c)) to reach the target pressure compared with a syringe-based system. Moreover, to demonstrate the performance with a large system interruption, we test the actuation system in one chamber which has air leakage.
It can be clearly found in Fig.~\ref{fig:mannequin}(c) that the syringe-based setup~\cite{mannequin_Syringe} cannot keep a constant pressure in the long term since no complement air flow is provided. Instead, our OpenPneu setup can always keep the target pressure with a steady state error at $0.2$ kPa, which is 0.3\% of the positive pressure range.

\subsection{Discussion}
While case studies presented in this section show the ability of our system to drive soft robots fabricated by silicon rubbers with multi-chambers, OpenPneu also has its limitation in performance -- such as the maximal pressure and the relative slow response on the exhaust process.
Since micro air pumps are selected in building a compact system, only limited airflow is provided to the system thus becoming less effective and responding when actuating a system requires high air pressure (i.e., higher than 2 bar). On the other hand, OpenPneu cannot precisely track the continuously decreased target pressure as discussed in Sec.~\ref{subsec:softhands}. This can be solved by replacing solenoid valves with expensive proportional piezo valves, however, will highly increase the cost of the system. The response time of our system is also affected by the volume of the chambers designed in soft robots. Those drawbacks could be solved by replacing main components based on application. The presented OpenPneu prototypes provide cost-effective modular solutions, which are easy to be reproduced and used by soft robot researchers. 

\section{Conclusions}
We present OpenPneu which is a highly-integrated pneumatic actuation system that effectively provides multiple pressure supplies. As self-contained equipment, it shows good compatibility with no requirement for compressed air as input. Modular design on the mechanical part and the electronic part make the system scalable 
and we demonstrate a prototype that contains ten air channels. Customized PCB and firmware are designed to effectively drive the system and achieve fast and stable closed-loop pressure control. We have demonstrated the capability of OpenPneu in soft robotic applications through three case studies, which all show good statics and dynamic performance. 
All the documents of OpenPneu are open-source to the researchers and we provide Python coding interface to make the system easy to use. 

\bibliographystyle{IEEEtran}
\bibliography{ICRA23OpenPneu}

\end{document}